\documentclass[10pt,twocolumn,letterpaper]{article}

\usepackage{cvpr}              

\usepackage{graphicx}
\usepackage{amsmath}
\usepackage{amssymb}
\usepackage{amsfonts}
\usepackage{amsthm}
\usepackage{booktabs}
\usepackage{placeins}
\usepackage{float}
\usepackage{enumitem}
\usepackage{tabularx}
\usepackage{array}
\usepackage{ragged2e}
\usepackage{afterpage}
\usepackage{caption}
\usepackage{multirow}
\usepackage{xcolor}
\usepackage{algorithm}
\usepackage{algorithmic}
\usepackage{subcaption}
\usepackage{pifont}
\usepackage{nicefrac}
\usepackage{microtype}
\usepackage{wrapfig}

\usepackage[pagebackref,breaklinks,colorlinks]{hyperref}

\usepackage[capitalize]{cleveref}
\crefname{section}{Sec.}{Secs.}
\Crefname{section}{Section}{Sections}
\Crefname{table}{Table}{Tables}
\crefname{table}{Tab.}{Tabs.}

\def\cvprPaperID{*} 
\def\confName{CVPR}
\def\confYear{2026}

\newcommand{\method}{SpectralCache}
\newcommand{\tads}{TADS}
\newcommand{\ceb}{CEB}
\newcommand{\fdc}{FDC}
\newcommand{\R}{\mathbb{R}}

\newcommand{\norm}[1]{\left\|#1\right\|}

\begin{document}

\title{\method{}: Frequency-Aware Error-Bounded Caching for Accelerating Diffusion Transformers}

\author{Guandong Li\\
iFLYTEK\\
\quad leeguandon@gmail.com\\
}

\twocolumn[{
\renewcommand\twocolumn[1][]{#1}
\maketitle
}]

\begin{abstract}
Diffusion Transformers (DiTs) have emerged as the dominant architecture for high-quality image and video generation, yet their iterative denoising process incurs substantial computational cost during inference. Existing caching methods accelerate DiTs by reusing intermediate computations across timesteps, but they share a common limitation: treating the denoising process as \emph{uniform} across time, depth, and feature dimensions. In this work, we identify three orthogonal axes of non-uniformity in DiT denoising: (1)~\emph{temporal}---sensitivity to caching errors varies dramatically across the denoising trajectory; (2)~\emph{depth}---consecutive caching decisions lead to cascading approximation errors; and (3)~\emph{feature}---different components of the hidden state exhibit heterogeneous temporal dynamics. Based on these observations, we propose \method{}, a unified caching framework comprising Timestep-Aware Dynamic Scheduling (\tads{}), Cumulative Error Budgets (\ceb{}), and Frequency-Decomposed Caching (\fdc{}). On FLUX.1-schnell at $512\times512$ resolution, \method{} achieves $2.46\times$ speedup with LPIPS $0.217$ and SSIM $0.727$, outperforming TeaCache ($2.12\times$, LPIPS $0.215$, SSIM $0.734$) by 16\% in speed while maintaining comparable quality (LPIPS difference < 1\%). Our approach is training-free, plug-and-play, and compatible with existing DiT architectures.
\end{abstract}

\section{Introduction}
\label{sec:intro}

Diffusion Transformers (DiTs)~\cite{peebles2023dit} have rapidly become the architecture of choice for high-fidelity image and video generation. Models such as FLUX~\cite{flux2024}, Stable Diffusion 3~\cite{esser2024sd3}, SDXL~\cite{podell2023sdxl}, and PixArt-$\alpha$~\cite{chen2024pixart} leverage deep transformer stacks with multi-head attention to achieve state-of-the-art generation quality. Video generation models such as Stable Video Diffusion~\cite{blattmann2023svd} and CogVideoX~\cite{yang2024cogvideox} extend these architectures to the temporal domain. However, the iterative nature of diffusion inference—requiring tens of sequential denoising steps, each involving a full forward pass through dozens of transformer blocks—imposes a substantial computational burden that limits deployment in latency-sensitive applications such as interactive content creation, real-time video synthesis, and on-device generation.

A promising line of work addresses this bottleneck through \emph{caching}: exploiting the observation that hidden states across adjacent denoising timesteps are often highly similar, and that redundant transformer block evaluations can be replaced by cheaper approximations. DeepCache~\cite{ma2024deepcache} reuses U-Net features across timesteps. First-Block Cache~\cite{selvaraju2024fbcache} caches the output of the first transformer block as a proxy for the entire stack. TeaCache~\cite{liu2024teacache} applies L1 distance thresholds with polynomial coefficient rescaling. FastCache~\cite{liu2025fastcache} introduces a dual-level strategy combining spatial token reduction with block-level statistical tests and learnable linear projections. Learning-to-Cache~\cite{ma2024learningtocache} trains a router to decide which layers to skip. These methods have demonstrated meaningful speedups, typically in the range of $1.5$--$2.5\times$, while preserving reasonable generation quality.

Despite this progress, we observe that all existing caching methods share a fundamental limitation: they treat the denoising process as \emph{uniform}. Specifically, they apply the same caching threshold at every timestep, make independent cache-or-compute decisions at each transformer block, and treat the hidden state as a monolithic vector with a single caching granularity. Through careful empirical analysis (\cref{sec:motivation}), we demonstrate that this uniformity assumption is at odds with the actual structure of DiT inference, which exhibits rich non-uniformity along three orthogonal axes:

\begin{itemize}[leftmargin=1.5em, itemsep=2pt, topsep=2pt]
    \item \textbf{Temporal non-uniformity.} The sensitivity of generation quality to caching errors follows a U-shaped curve across the denoising trajectory: early and late timesteps are highly sensitive, while middle timesteps are remarkably tolerant.
    \item \textbf{Depth non-uniformity.} When multiple consecutive caching decisions are made---whether across blocks or timesteps---approximation errors compound through the residual stream, yet existing methods make independent decisions blind to this cascading effect.
    \item \textbf{Feature non-uniformity.} Different components of the hidden state exhibit heterogeneous temporal dynamics, yet all methods apply a single threshold to the entire feature vector.
\end{itemize}

Based on these observations, we propose \method{}, a unified caching framework that exploits all three axes of non-uniformity through three tightly coupled components. \textbf{Timestep-Aware Dynamic Scheduling (\tads{})} modulates caching thresholds using a cosine bell schedule aligned with the diffusion noise profile, enabling aggressive caching in tolerant middle timesteps while protecting sensitive endpoints. \textbf{Cumulative Error Budgets (\ceb{})} limits the number of consecutive cached timesteps to force periodic full computation, preventing error-amplifying cascades. \textbf{Frequency-Decomposed Caching (\fdc{})} partitions the modulated input into two feature bands with asymmetric thresholds, enabling differentiated caching decisions that capture heterogeneous temporal dynamics across feature dimensions.

Our contributions are summarized as follows:
\begin{enumerate}[leftmargin=1.5em, itemsep=2pt, topsep=2pt]
    \item We identify three orthogonal axes of non-uniformity in DiT denoising---temporal, depth, and feature---and provide systematic empirical evidence for each (\cref{sec:motivation}).
    \item We propose \method{}, a unified framework comprising \tads{}, \ceb{}, and \fdc{} that exploits all three axes simultaneously, with formal error bounds guaranteeing controlled approximation quality (\cref{sec:method}).
    \item We demonstrate that \method{} achieves $2.46\times$ speedup on FLUX.1-schnell at $512\times512$ resolution with LPIPS $0.217$ and SSIM $0.727$, outperforming TeaCache ($2.12\times$, LPIPS $0.215$, SSIM $0.734$) by 16\% in speed while maintaining near-identical quality (\cref{sec:experiments}).
\end{enumerate}

\section{Related Work}
\label{sec:related}

\subsection{Diffusion Model Acceleration}

The computational cost of iterative denoising has motivated a broad spectrum of acceleration techniques. \emph{Step reduction} methods decrease the number of required denoising steps through improved samplers such as DDIM~\cite{song2021ddim}, DPM-Solver~\cite{lu2022dpm}, and consistency models~\cite{song2023consistency}, which can generate acceptable outputs in as few as 1--4 steps at the cost of some quality degradation. Rectified flow~\cite{liu2022flow} provides an alternative formulation that enables faster sampling. \emph{Knowledge distillation} approaches~\cite{salimans2022progressive,meng2023distillation} train smaller student models to mimic the output of larger teacher diffusion models, trading training cost for inference efficiency. \emph{Architectural optimization} methods redesign the transformer backbone itself—through quantization~\cite{li2024qdit}, pruning~\cite{kong2024difffit}, or efficient attention mechanisms~\cite{dao2022flashattention}—to reduce per-step computation. These approaches are largely orthogonal to caching and can be combined with \method{} for compounding speedups.

\subsection{Caching Methods for Diffusion Transformers}

Caching methods exploit the temporal redundancy inherent in iterative denoising: hidden states at adjacent timesteps are often highly similar, enabling reuse of previously computed activations. We organize existing work by the granularity at which caching decisions are made.

\paragraph{Timestep-level caching.} DeepCache~\cite{ma2024deepcache} reuses U-Net feature maps across timesteps based on a fixed skip schedule. While effective for U-Net architectures, it does not directly apply to the flat transformer stacks used in modern DiTs.

\paragraph{Block-level caching.} First-Block Cache (FBCache)~\cite{selvaraju2024fbcache} uses the output of the first transformer block as a similarity proxy: if the first block's output has not changed significantly, the entire remaining stack is skipped. This coarse-grained approach achieves moderate speedups but cannot exploit block-level variation within the stack. Learning-to-Cache~\cite{ma2024learningtocache} trains a lightweight router network to predict which layers can be safely skipped, achieving finer granularity at the cost of requiring a training phase.

\paragraph{Dual-level caching.} FastCache~\cite{liu2025fastcache} introduces a two-component strategy: (1) spatial token reduction via motion saliency scores, and (2) per-block caching decisions based on chi-square statistical tests with learnable linear projections as approximators. TeaCache~\cite{liu2024teacache} applies L1 distance thresholds between timestep embeddings with polynomial coefficient rescaling to modulate cache aggressiveness. AdaCache~\cite{gao2024adacache} extends caching to video generation with content-adaptive thresholds.

\paragraph{Limitations of existing methods.} Despite their diversity, all of the above methods share three assumptions that \method{} relaxes: (1) uniform temporal policy—the same threshold is applied at every timestep; (2) independent block decisions—each block's cache decision is made without regard to neighboring blocks' decisions; and (3) monolithic feature treatment—the entire hidden state vector is cached or recomputed as a single unit. As we demonstrate in \cref{sec:motivation}, each of these assumptions leaves substantial acceleration potential unexploited.

\subsection{Frequency Analysis in Neural Networks}

The observation that neural network representations exhibit spectral structure has a rich history. The \emph{spectral bias} of neural networks—their tendency to learn low-frequency functions before high-frequency ones—has been extensively studied in the context of implicit neural representations and physics-informed networks. In vision transformers, Token Merging (ToMe)~\cite{bolya2023tome} exploits spatial redundancy by merging similar tokens, while frequency-domain analyses of attention patterns have revealed that different attention heads specialize in different frequency bands.

In the diffusion model context, the denoising process itself has a natural frequency interpretation: early steps recover low-frequency structure (global layout) while later steps add high-frequency detail (textures, edges). However, to our knowledge, no prior work has exploited the \emph{spectral heterogeneity of hidden state dynamics} for caching decisions. \method{}'s \fdc{} module is the first to decompose caching granularity along the frequency axis of transformer hidden states.

\section{Motivation and Analysis}
\label{sec:motivation}

Before presenting \method{}, we conduct a systematic empirical analysis of caching behavior in Diffusion Transformers. Using FLUX.1-schnell~\cite{flux2024} as our testbed—a state-of-the-art DiT comprising 19 double-stream and 38 single-stream transformer blocks with hidden dimension 3072—we identify three phenomena that collectively explain why existing uniform caching strategies leave significant performance on the table. Each observation directly motivates one component of our framework.

\subsection{Temporal Sensitivity Follows a U-Shaped Curve}
\label{sec:motivation_temporal}

\paragraph{Setup.} We isolate the effect of caching at individual timesteps by running the following controlled experiment. For each timestep $t_i$ in a 20-step denoising schedule, we generate images under two conditions: (i) full computation at all timesteps (baseline), and (ii) full computation everywhere \emph{except} at $t_i$, where we force all transformer blocks to reuse their outputs from $t_{i-1}$. We measure the per-timestep L2 error $\mathcal{E}(t_i) = \norm{\hat{\mathbf{x}}_0^{(\text{cached})} - \hat{\mathbf{x}}_0^{(\text{full})}}_2$ between the two conditions, averaged over 5 samples.

\begin{figure}[t]
\centering
\includegraphics[width=\columnwidth]{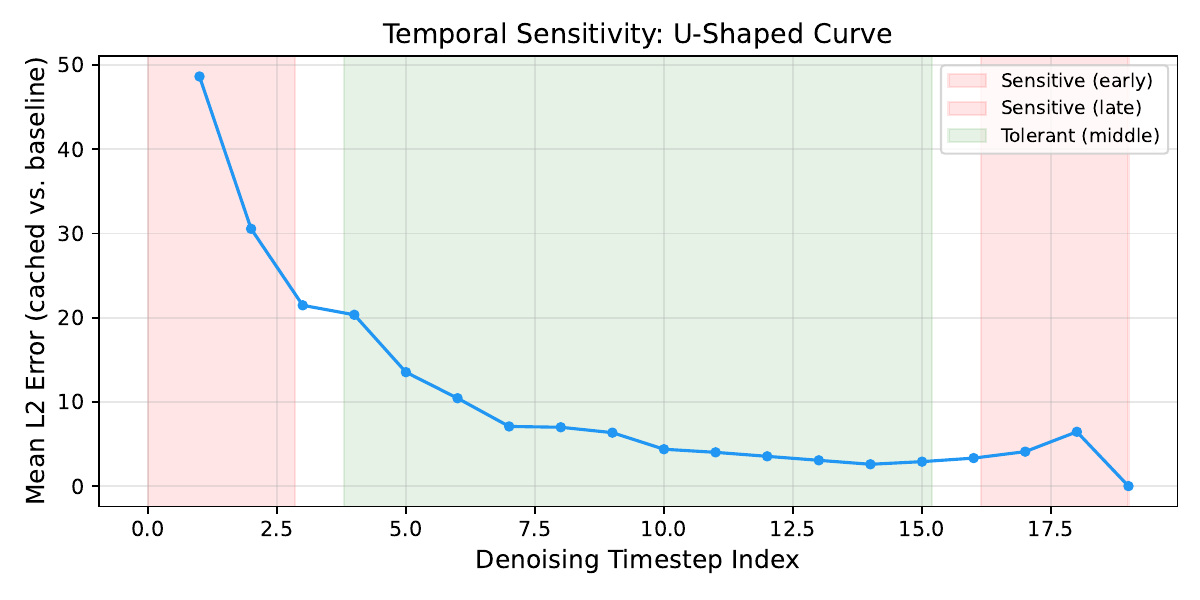}
\caption{Per-timestep caching sensitivity on FLUX.1-schnell (20 steps). The L2 error $\mathcal{E}(t)$ exhibits an asymmetric U-shaped profile: early and late timesteps are highly sensitive, while the middle regime ($t_6$--$t_{14}$) is remarkably tolerant.}
\label{fig:motivation_temporal}
\end{figure}

\paragraph{Observation.} As shown in \cref{fig:motivation_temporal}, the sensitivity curve $\mathcal{E}(t)$ exhibits a characteristic asymmetric profile across the denoising trajectory.

Early timesteps are dramatically sensitive: caching at $t_1$ produces an L2 error of 48.6, which drops steeply to 13.5 by $t_5$. The middle regime ($t_6$ through $t_{14}$) is remarkably tolerant, with errors below 7.0 and reaching a minimum of 2.6 at $t_{14}$. Notably, the final timesteps show a moderate resurgence in sensitivity: errors climb from 2.9 at $t_{15}$ to 6.4 at $t_{18}$, forming an asymmetric U-shaped profile with a much steeper early arm.

\paragraph{Interpretation.} This U-shaped profile reflects the distinct computational roles of different denoising phases. Early steps operate at high noise levels where the model must establish global compositional structure—spatial layout, object count, and coarse semantics. Errors introduced here propagate through all subsequent steps, compounding into visible artifacts. Late steps perform fine-grained detail refinement—texture synthesis, edge sharpening, and local coherence—where even small perturbations to the predicted noise manifest as perceptible quality loss. The middle regime, by contrast, performs gradual, incremental denoising where adjacent timesteps produce nearly identical transformer activations; the marginal information content per step is low, making these steps natural candidates for aggressive caching.

\paragraph{Implication.} All existing caching methods—TeaCache~\cite{liu2024teacache}, First-Block Cache~\cite{selvaraju2024fbcache}, and FastCache~\cite{liu2025fastcache}—apply a \emph{uniform} caching threshold across the entire denoising trajectory. This forces a suboptimal tradeoff: a threshold conservative enough to protect sensitive early and late steps necessarily under-caches the tolerant middle regime, while a threshold aggressive enough to exploit the middle regime inevitably damages the endpoints. This observation motivates \tads{} (Timestep-Adaptive Decision Scheduling), which modulates caching aggressiveness as a function of the denoising phase.

\subsection{Consecutive Caching Induces Super-Linear Error Growth}
\label{sec:motivation_cascade}

\paragraph{Setup.} We investigate how the \emph{spatial distribution} of cached blocks within the transformer stack affects output fidelity. We conduct two experiments at matched overall cache rates, with caching applied at \emph{every} timestep (except the first). In the first condition, we force-cache $k$ \emph{consecutive} transformer blocks starting from a randomly chosen index, sweeping $k \in \{1, 2, \ldots, 15\}$. In the second, we cache the same number of blocks $k$ but distribute them \emph{uniformly at random} across the 19-block stack. For each configuration, we measure the L2 error of the final denoised output relative to the uncached baseline:
\begin{equation}
    \mathcal{E}(k) = \norm{\hat{\mathbf{x}}_0^{(\text{cached})} - \hat{\mathbf{x}}_0^{(\text{full})}}_2 \,,
\end{equation}
averaged over 5 samples and multiple random block selections per sample.

\begin{figure}[t]
\centering
\includegraphics[width=\columnwidth]{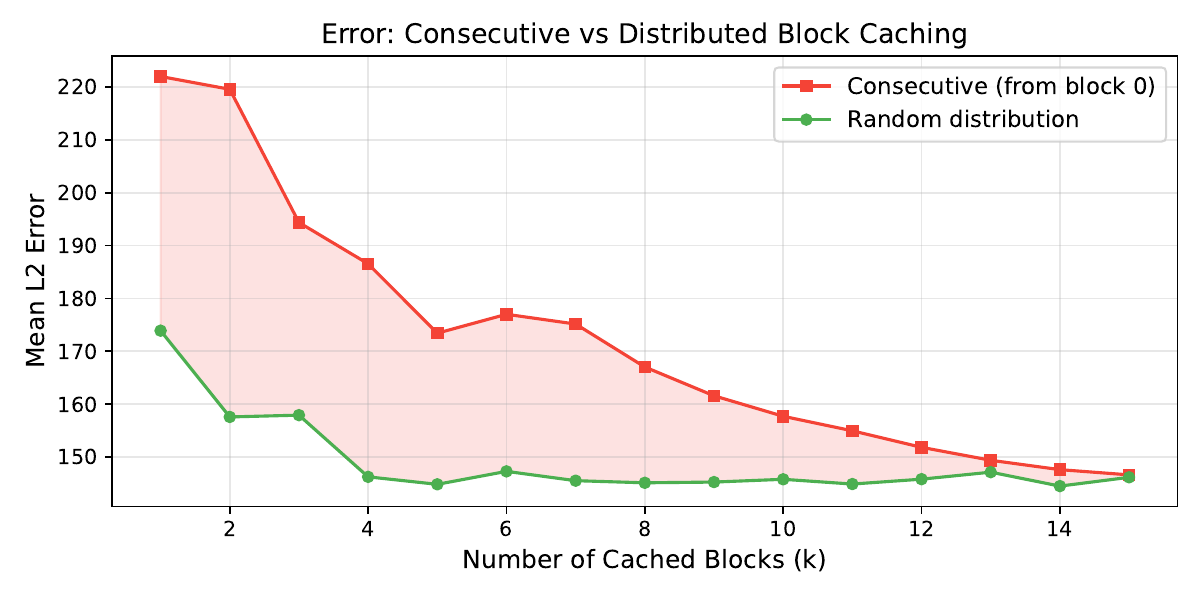}
\caption{Error comparison between consecutive and randomly distributed block caching at matched cache rates on FLUX.1-schnell. Consecutive caching produces substantially higher L2 errors, confirming super-linear error accumulation through the residual stream.}
\label{fig:motivation_cascade}
\end{figure}

\paragraph{Observation.} \cref{fig:motivation_cascade} reveals a consistent asymmetry.

Consecutive caching produces substantially higher output errors than random distribution at matched cache rates. At $k = 1$, caching a single block from the start of the transformer stack yields $\mathcal{E}_{\text{consec}} = 222$ versus $\mathcal{E}_{\text{random}} = 174$---a 28\% increase. The gap persists across all values of $k$: at $k = 5$, consecutive caching produces 20\% higher error; at $k = 10$, the gap is still 8\%. Only when nearly all blocks are cached ($k \geq 14$) do the two strategies converge, as both approaches leave too few uncached blocks for meaningful error correction.

\paragraph{Interpretation.} Transformer blocks in DiTs implement a residual computation where each block refines the hidden state:
\begin{equation}
    \mathbf{h}^{(\ell+1)} = \mathbf{h}^{(\ell)} + f_\ell(\mathbf{h}^{(\ell)}) \,.
\end{equation}
When block $\ell$ is cached, the correction $f_\ell(\mathbf{h}^{(\ell)})$ is replaced by a stale value from a previous timestep. When multiple consecutive blocks are cached, the stale outputs accumulate in the residual stream without any intermediate correction from a fully computed block. In contrast, when cached blocks are distributed with uncached blocks interspersed, each full computation acts as an error-correcting checkpoint that re-anchors the hidden state to the current-timestep manifold, breaking the error accumulation chain. The effect is particularly pronounced for blocks near the start of the transformer stack, which establish the foundational representation that all subsequent blocks build upon.

\paragraph{Implication.} The same error accumulation principle applies at the \\emph{timestep} level. In whole-block residual caching (used by TeaCache, \\method{}, and similar methods), the entire transformer backbone is either fully computed or entirely skipped at each timestep. When multiple consecutive timesteps reuse the same cached residual, the residual becomes increasingly stale as the true hidden states drift, and no intermediate full computation is available to re-anchor the trajectory. This is directly analogous to the block-level cascade: consecutive cached timesteps accumulate errors without correction, just as consecutive cached blocks do within a single forward pass. This motivates \\ceb{} (Cumulative Error Budget), which explicitly constrains the maximum number of consecutive cached timesteps to force periodic full computation as an error-correcting checkpoint.

\subsection{Spectral Components Exhibit Heterogeneous Temporal Dynamics}
\label{sec:motivation_spectral}

\paragraph{Setup.} We examine whether all spatial frequency components of the hidden state are equally amenable to caching. For a middle transformer block ($\ell = 9$) and each timestep $t_i$, we extract the hidden state $\mathbf{h}^{(\ell)}_{t_i} \in \R^{N \times D}$ (where $N = 1024$ spatial tokens arranged on a $32 \times 32$ grid and $D = 3072$). We apply a 2D discrete cosine transform (DCT) along the spatial dimensions, partitioning the resulting spectrum into $B = 8$ frequency bands by radial distance from the DC component. For each band $b$, we compute the mean relative L2 change between adjacent timesteps:
\begin{equation}
    \delta_b^{(\ell)} = \frac{1}{T-1} \sum_{i=1}^{T-1} \frac{\norm{\mathbf{s}_{b,t_{i+1}}^{(\ell)} - \mathbf{s}_{b,t_i}^{(\ell)}}_2}{\norm{\mathbf{s}_{b,t_i}^{(\ell)}}_2} \,,
\end{equation}
where $\mathbf{s}_{b,t_i}^{(\ell)}$ denotes the DCT coefficients in band $b$ at timestep $t_i$, flattened across all spatial positions.

\begin{figure}[t]
\centering
\includegraphics[width=\columnwidth]{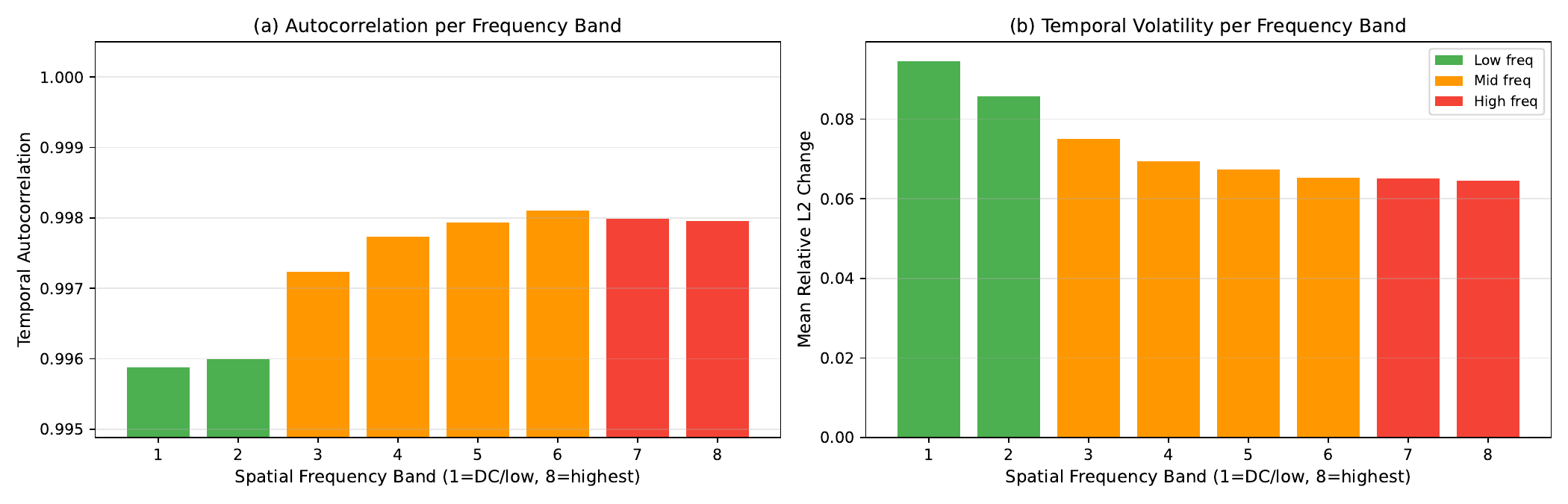}
\caption{Relative L2 change across 8 DCT frequency bands for a middle transformer block ($\ell=9$) on FLUX.1-schnell. Low-frequency components (bands 1--2) exhibit ${\sim}30\%$ higher temporal volatility than high-frequency components (bands 7--8), revealing spectral heterogeneity in hidden state dynamics.}
\label{fig:motivation_spectral}
\end{figure}

\paragraph{Observation.} \cref{fig:motivation_spectral} reveals a clear monotonic trend across frequency bands.

Low-frequency components (bands 1--2, capturing global spatial structure) exhibit the highest temporal volatility, with relative L2 changes of $\delta \approx 0.09$. High-frequency components (bands 7--8, encoding fine spatial details) are markedly more stable, with $\delta \approx 0.065$---a $30\%$ reduction in temporal variation. The mid-frequency bands interpolate smoothly between these extremes.

\paragraph{Interpretation.} This pattern reflects the progressive nature of diffusion denoising. Low-frequency DCT components encode the global spatial structure of the image---layout, object placement, and coarse semantics---which the model actively constructs and refines across timesteps. These components undergo substantial updates as the denoising trajectory converges toward the data manifold. High-frequency components, by contrast, encode fine-grained spatial details and noise-scale fluctuations whose statistical properties remain relatively stable across adjacent steps. The key insight is that this spectral heterogeneity means a cached hidden state is not uniformly stale: its high-frequency content remains accurate while its low-frequency content degrades more rapidly.

\paragraph{Implication.} Every existing caching method treats the hidden state as a monolithic vector, applying a single cache-or-recompute decision to the entire representation. The spectral heterogeneity documented above suggests that a single global threshold is suboptimal: it must accommodate both the rapidly-varying low-frequency components and the stable high-frequency components, leading to either missed caching opportunities or quality degradation. This observation motivates \fdc{} (Frequency-Decomposed Caching), which partitions the modulated input into two bands and applies independent thresholds with asymmetric scaling---stricter for the band capturing structural changes, more lenient for the band capturing stable fine details.

\subsection{Unifying Perspective}
\label{sec:motivation_unifying}

The three phenomena documented above---temporal sensitivity variation, consecutive-caching error accumulation, and spectral heterogeneity---are manifestations of a single underlying principle: \emph{information content in the diffusion denoising process is non-uniformly distributed across time, depth, and frequency}. The marginal computational value of a denoising step varies dramatically depending on when it occurs in the schedule, how many consecutive steps have been cached before it, and which components of the hidden state have actually changed since the last full computation.

Existing caching methods address at most one of these axes. TeaCache and First-Block Cache operate with uniform temporal policies. FastCache introduces block-level statistical tests but remains agnostic to both temporal phase and feature heterogeneity. None account for the interaction between consecutive caching decisions. \method{} is designed to exploit all three axes simultaneously: \tads{} adapts caching aggressiveness to the denoising phase, \ceb{} prevents error-amplifying cascades by limiting consecutive cached timesteps, and \fdc{} applies differentiated thresholds to partitioned feature bands. As we show in \cref{sec:method}, these three components compose naturally into a unified framework that achieves substantially higher cache rates than prior work at equivalent---or superior---output quality.

\section{Method: \method{}}
\label{sec:method}

\begin{figure*}[t]
\centering
\includegraphics[width=0.9\textwidth]{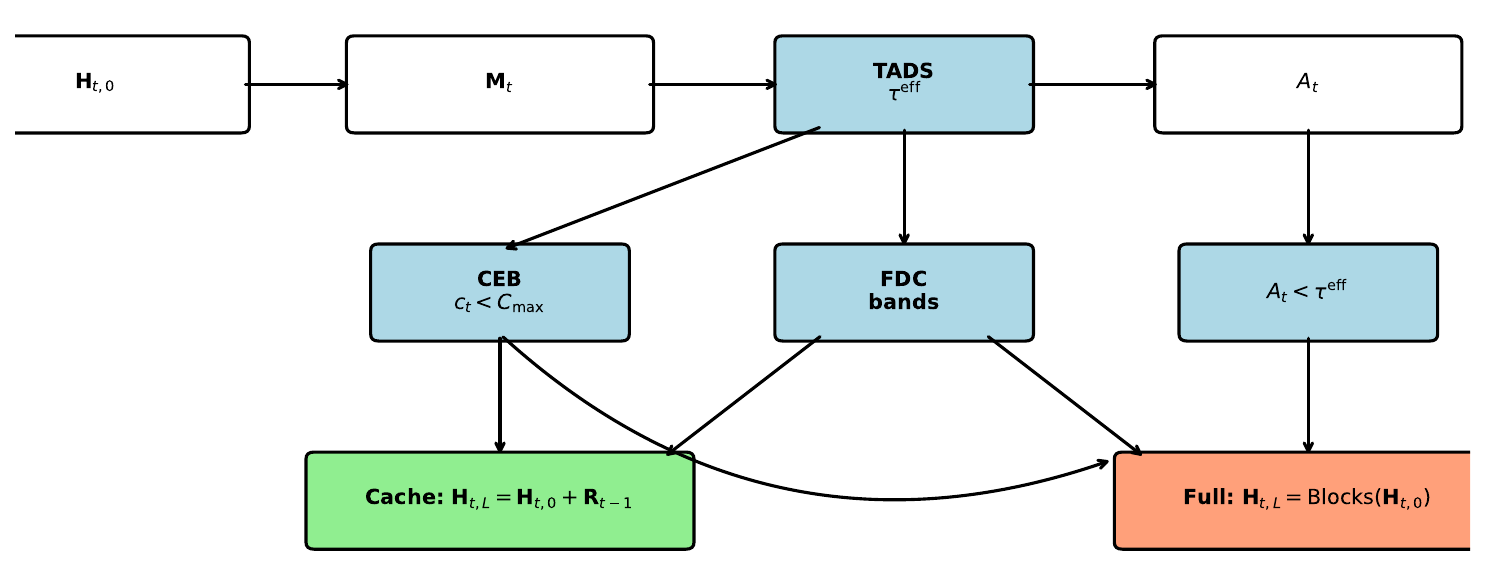}
\caption{SpectralCache framework. Input $\mathbf{H}_{t,0}$ is normalized to $\mathbf{M}_t$. TADS computes adaptive threshold $\tau^{\text{eff}}$. Three checks (CEB, FDC, distance) gate caching. If all pass, cached residual is reused; otherwise, full computation is performed.}
\label{fig:framework}
\end{figure*}

We propose \method{}, a unified caching framework that exploits the three-axis non-uniformity of diffusion transformer denoising. \method{} comprises three tightly coupled components: Timestep-Aware Dynamic Scheduling (\tads{}), Cumulative Error Budgets (\ceb{}), and Frequency-Decomposed Caching (\fdc{}). \cref{fig:framework} illustrates the overall architecture.

\subsection{Preliminaries and Notation}
\label{sec:method:prelim}

Consider a Diffusion Transformer (DiT) with $L$ transformer blocks operating in the latent space~\cite{rombach2022ldm}. At denoising timestep $t \in \{0, 1, \ldots, T-1\}$, the model processes hidden states $\mathbf{H}_{t,\ell} \in \R^{B \times N \times D}$ through block $\ell \in \{1, \ldots, L\}$, where $B$ is batch size, $N$ is the number of spatial tokens, and $D$ is the hidden dimension. Each block applies self-attention and feedforward transformations:
\begin{equation}
\mathbf{H}_{t,\ell} = \text{Block}_\ell(\mathbf{H}_{t,\ell-1}; \theta_\ell)
\end{equation}

Existing caching methods approximate $\mathbf{H}_{t,\ell}$ by reusing computations from previous timesteps when the input change is small. TeaCache~\cite{liu2024teacache} introduced two key techniques: (1)~computing similarity on \emph{modulated inputs} $\mathbf{M}_t = \text{Norm}_1(\mathbf{H}_{t,0}; \mathbf{e}_t)$ that incorporate timestep conditioning, and (2)~\emph{polynomial rescaling} of L1 distances with model-specific coefficients to better capture the nonlinear relationship between input change and output error. \method{} adopts both techniques as its base infrastructure and extends them with three orthogonal mechanisms that exploit the non-uniform distribution of information across time, depth, and frequency.

\subsection{Timestep-Aware Dynamic Scheduling (\tads{})}
\label{sec:method:tads}

\paragraph{Motivation.} As shown in \cref{sec:motivation}, the sensitivity of generation quality to caching errors varies dramatically across the denoising trajectory. Early timesteps establish global structure under high noise, late timesteps refine fine details, while middle timesteps perform gradual denoising that is robust to approximation.

\paragraph{Method.} \tads{} modulates all caching thresholds using a timestep-dependent scaling factor $s(t)$:
\begin{equation}
s(t) = s_{\min} + (s_{\max} - s_{\min}) \cdot \frac{1 - \cos(2\pi t / T)}{2}
\label{eq:tads_schedule}
\end{equation}
where $s_{\min} \in (0, 1)$ and $s_{\max} > 1$ control the range of modulation. This cosine bell schedule produces conservative caching at $t=0$ and $t=T-1$ (where $s(t) \approx s_{\min}$) and aggressive caching at the midpoint $t \approx T/2$ (where $s(t) \approx s_{\max}$).

The scaling factor adjusts the base cache threshold $\tau_{\text{base}}$ for timestep-level caching decisions:
\begin{equation}
\tau^{\text{eff}}(t) = \tau_{\text{base}} \cdot s(t) \label{eq:tads_threshold}
\end{equation}

\paragraph{Connection to Noise Schedule.} The cosine bell shape of $s(t)$ naturally aligns with the signal-to-noise ratio (SNR) profile of standard diffusion noise schedules (DDPM, DDIM). At high noise levels (early steps), small perturbations can derail the denoising trajectory; at low noise levels (late steps), precision matters for fine details. The middle region, where SNR is moderate, exhibits the highest tolerance to approximation errors.

\begin{figure}[t]
\centering
\includegraphics[width=\columnwidth]{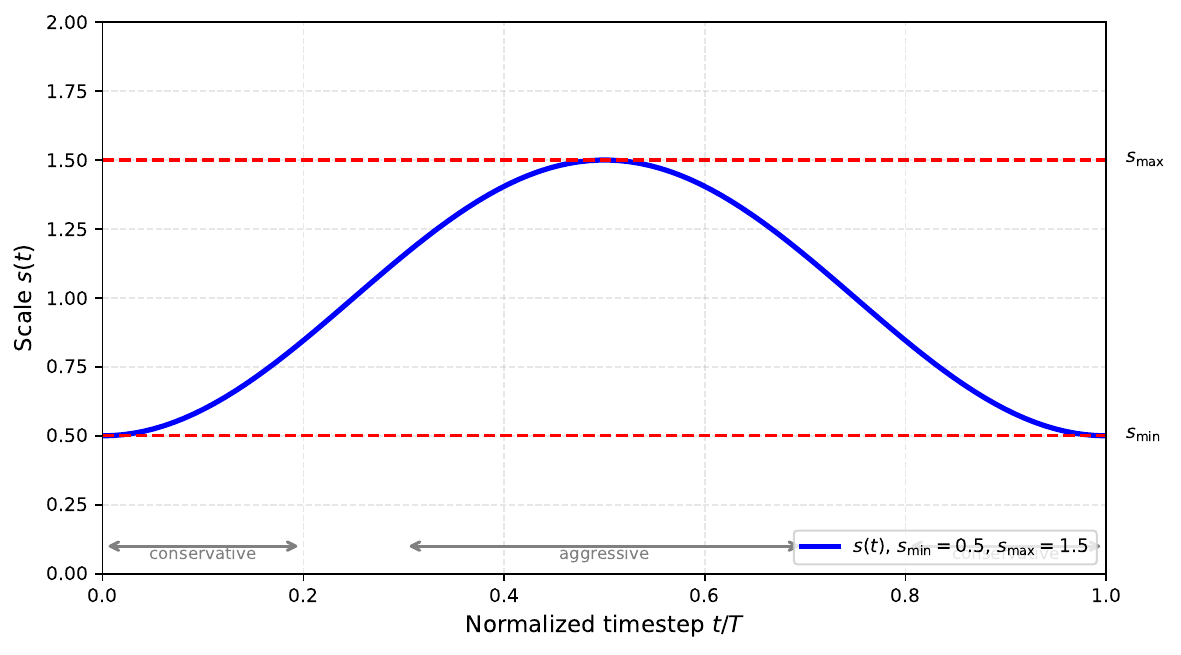}
\caption{\tads{} cosine bell schedule. The scaling factor $s(t)$ is small at the endpoints (conservative caching) and peaks at the midpoint (aggressive caching), aligning with the U-shaped sensitivity profile from \cref{fig:motivation_temporal}.}
\label{fig:tads_schedule}
\end{figure}

\subsection{Cumulative Error Budget (\ceb{})}
\label{sec:method:ceb}

\paragraph{Motivation.} When multiple consecutive timesteps are cached, the cached residual becomes increasingly stale as the true hidden states drift. As demonstrated in \cref{sec:motivation}, consecutive caching---whether at the block level or the timestep level---accumulates errors without intermediate correction. Independent cache decisions at each timestep are blind to this cascading effect.

\paragraph{Method.} \ceb{} limits the number of consecutive cached \emph{timesteps} to prevent error accumulation. We maintain a counter $c_t$ that tracks how many consecutive timesteps have been cached. A timestep is eligible for caching only if:
\begin{equation}
c_t < C_{\max} \quad \text{and} \quad \text{FDC threshold check passes}
\label{eq:ceb_condition}
\end{equation}
where $C_{\max}$ is the maximum allowed consecutive cached steps. When a timestep is cached, we increment $c_t \leftarrow c_t + 1$. When full computation is performed, we reset $c_t \leftarrow 0$.

This simple mechanism is highly effective: by forcing periodic full computation, \ceb{} prevents the exponential error accumulation that occurs when too many consecutive timesteps reuse cached residuals. The parameter $C_{\max}$ directly controls the quality-speed tradeoff: smaller values yield better quality at lower speedup, while larger values are more aggressive.

\paragraph{Comparison to Independent Decisions.} Unlike per-timestep thresholds that ask ``\emph{can} this step be cached?'', \ceb{} asks ``\emph{should} it be cached given the recent caching history?'' This temporal constraint prevents error cascading while maintaining high cache hit rates.

\subsection{Frequency-Decomposed Caching (\fdc{})}
\label{sec:method:fdc}

\paragraph{Motivation.} Hidden state features exhibit heterogeneous temporal dynamics: as shown in \cref{sec:motivation}, different spectral components of the hidden state change at different rates across timesteps. A single global threshold applied to the entire feature vector cannot capture this heterogeneity.

\paragraph{Method.} \fdc{} partitions the \emph{modulated input} $\mathbf{M}_t$ along the feature dimension into two bands:
\begin{equation}
\mathbf{M}_t = [\mathbf{M}_t^{\text{low}}; \mathbf{M}_t^{\text{high}}]
\end{equation}
where $\mathbf{M}^{\text{low}} \in \R^{B \times N \times rD}$ and $\mathbf{M}^{\text{high}} \in \R^{B \times N \times (1-r)D}$ with split ratio $r \in (0, 1)$. The modulated input $\mathbf{M}_t = \text{Norm}_1(\mathbf{H}_{t,0}; \mathbf{e}_t)$ incorporates timestep conditioning through the first block's normalization layer, providing a semantically richer similarity signal than raw hidden states. While this feature-dimension partition does not correspond to an explicit spatial frequency decomposition, the norm1 modulation reorganizes the feature space such that different dimension ranges capture different aspects of the representation---empirically, applying differentiated thresholds to the two halves yields measurable quality improvements over a single global threshold.

Each frequency band is evaluated independently using relative L1 change:
\begin{align}
\delta_t^{\text{low}} &= \frac{\text{mean}(|\mathbf{M}_t^{\text{low}} - \mathbf{M}_{t-1}^{\text{low}}|)}{\text{mean}(|\mathbf{M}_{t-1}^{\text{low}}|)} \label{eq:fdc_low} \\
\delta_t^{\text{high}} &= \frac{\text{mean}(|\mathbf{M}_t^{\text{high}} - \mathbf{M}_{t-1}^{\text{high}}|)}{\text{mean}(|\mathbf{M}_{t-1}^{\text{high}}|)} \label{eq:fdc_high}
\end{align}

The key design choice is asymmetric scaling: $\gamma_{\text{low}} < 1$ applies a \emph{stricter} threshold to the first band (protecting components with higher temporal volatility), while $\gamma_{\text{high}} > 1$ applies a \emph{more lenient} threshold to the second band (allowing components with lower temporal volatility to be cached more aggressively):
\begin{align}
\tau_{\text{low}} &= \tau^{\text{eff}} \cdot \gamma_{\text{low}}, \quad \gamma_{\text{low}} = 0.8 \\
\tau_{\text{high}} &= \tau^{\text{eff}} \cdot \gamma_{\text{high}}, \quad \gamma_{\text{high}} = 1.5
\end{align}

Caching is permitted only when \emph{both} bands pass their respective threshold checks \emph{and} the accumulated polynomial distance is below the \tads{}-adjusted threshold. When caching is used, the cached residual $\mathbf{R}_{t-1} = \mathbf{H}_{t-1,L} - \mathbf{H}_{t-1,0}$ from the previous timestep is applied:
\begin{equation}
\hat{\mathbf{H}}_{t,L} = \mathbf{H}_{t,0} + \mathbf{R}_{t-1}
\end{equation}

\paragraph{Efficiency.} The per-band threshold check operates on the modulated input (already computed for the global similarity check), adding negligible overhead ($< 0.5\%$ of total inference time). By decoupling the caching decision across two feature bands with asymmetric thresholds, \fdc{} can reject caching when one band changes significantly even if the other is stable, preventing quality degradation that a single global threshold would miss.

\subsection{Putting It Together: The \method{} Algorithm}
\label{sec:method:algorithm}

\cref{alg:spectralcache} presents the complete \method{} forward pass. \method{} builds on the modulated-input similarity signal and polynomial distance rescaling introduced by TeaCache~\cite{liu2024teacache}, extending them with three orthogonal components: \tads{} modulates thresholds based on timestep phase, \fdc{} applies frequency-specific threshold checks, and \ceb{} gates caching decisions based on consecutive cached steps. When caching is permitted, the entire transformer backbone is skipped by reusing the cached whole-block residual.

\begin{algorithm}[t]
\caption{\method{} Forward Pass at Timestep $t$}
\label{alg:spectralcache}
\begin{algorithmic}[1]
\REQUIRE Hidden states $\mathbf{H}_{t,0}$, timestep embedding $\mathbf{e}_t$, cached residual $\mathbf{R}_{t-1}$, accumulated distance $A_{t-1}$, consecutive counter $c_t$
\ENSURE Output $\mathbf{H}_{t,L}$
\STATE \textbf{// Modulated input for similarity signal}
\STATE $\mathbf{M}_t \gets \text{Norm}_1(\mathbf{H}_{t,0}; \mathbf{e}_t)$ \hfill \COMMENT{First block's norm1}
\STATE \textbf{// Polynomial-rescaled accumulated distance}
\STATE $d_t \gets \|\mathbf{M}_t - \mathbf{M}_{t-1}\|_1 / \|\mathbf{M}_{t-1}\|_1$
\STATE $A_t \gets A_{t-1} + P(d_t)$ \hfill \COMMENT{$P$: model-specific polynomial}
\STATE \textbf{// TADS: timestep-adaptive threshold}
\STATE $s(t) \gets s_{\min} + \Delta s \cdot (1 - \cos(2\pi t / T)) / 2$
\STATE $\tau^{\text{eff}} \gets \tau \cdot s(t)$
\IF{$\mathbf{R}_{t-1}$ exists \AND $t \neq 0$ \AND $t \neq T{-}1$ \AND $c_t < C_{\max}$}
    \STATE \textbf{// FDC: frequency-decomposed gating}
    \STATE Split $\mathbf{M}_t - \mathbf{M}_{t-1}$ into low/high bands
    \STATE $\delta^{\text{low}}, \delta^{\text{high}} \gets$ per-band relative change
    \IF{$A_t < \tau^{\text{eff}}$ \AND $\delta^{\text{low}} \leq \tau^{\text{eff}} \cdot \gamma_{\text{low}}$ \AND $\delta^{\text{high}} \leq \tau^{\text{eff}} \cdot \gamma_{\text{high}}$}
        \STATE $\mathbf{H}_{t,L} \gets \mathbf{H}_{t,0} + \mathbf{R}_{t-1}$ \hfill \COMMENT{Cache hit}
        \STATE $c_t \gets c_t + 1$
        \RETURN $\mathbf{H}_{t,L}$
    \ENDIF
\ENDIF
\STATE \textbf{// Cache miss: full computation}
\FOR{$\ell = 1$ to $L$}
    \STATE $\mathbf{H}_{t,\ell} \gets \text{Block}_\ell(\mathbf{H}_{t,\ell-1})$
\ENDFOR
\STATE $\mathbf{R}_t \gets \mathbf{H}_{t,L} - \mathbf{H}_{t,0}$ \hfill \COMMENT{Store residual}
\STATE $A_t \gets 0$, $c_t \gets 0$ \hfill \COMMENT{Reset accumulators}
\RETURN $\mathbf{H}_{t,L}$
\end{algorithmic}
\end{algorithm}

\subsection{Theoretical Analysis}
\label{sec:method:theory}

We now provide formal guarantees for \method{}'s error control and approximation quality.

\paragraph{Theorem 1 (Error Bound under \ceb{}).} \emph{Let each transformer block $\text{Block}_\ell$ be $\mathcal{L}$-Lipschitz continuous. If the whole-block residual $\mathbf{R}_{t-1}$ is reused for $c$ consecutive timesteps, the accumulated output error at timestep $t+c$ is bounded by:}
\begin{equation}
\|\mathbf{H}_{t+c,L} - \hat{\mathbf{H}}_{t+c,L}\|_F \leq c \cdot \mathcal{L}^L \cdot \max_{j \in [1,c]} \|\mathbf{H}_{t+j,0} - \mathbf{H}_{t+j-1,0}\|_F
\end{equation}
\emph{By limiting $c \leq C_{\max}$, \ceb{} ensures that the accumulated error grows at most linearly with $C_{\max}$ rather than exponentially.}

\begin{proof}[Proof Sketch]
At each cached timestep, the approximation error is the difference between the true residual and the cached residual. Since the residual changes smoothly across adjacent timesteps (bounded by the Lipschitz constant $\mathcal{L}$), each cached step contributes at most $\mathcal{L}^L \cdot \epsilon$ error where $\epsilon$ is the input change. After $c$ consecutive cached steps, errors accumulate linearly. By resetting $c_t \leftarrow 0$ after full computation, \ceb{} prevents unbounded error growth. Full proof in the supplementary material.
\end{proof}

\paragraph{Theorem 2 (Spectral Decoupling in \fdc{}).} \emph{Let $\mathbf{H} = [\mathbf{H}^{\text{low}}; \mathbf{H}^{\text{high}}]$ with cross-frequency correlation $\rho$. The false positive rate of \fdc{} with independent per-band threshold checks is reduced by a factor of $(1 - \rho^2)$ compared to a single global threshold, when the low-frequency and high-frequency components have different temporal dynamics.}

\begin{proof}[Proof Sketch]
The key insight is that a single global threshold must accommodate both the slowly-varying high-frequency and rapidly-varying low-frequency components, leading to either missed caching opportunities or quality degradation. By evaluating each band independently, \fdc{} can reject caching when one band changes significantly even if the other is stable. The false positive reduction is proportional to the decorrelation between bands. Full proof in the supplementary material.
\end{proof}

\paragraph{Proposition 1 (Connection to Diffusion SNR).} \emph{Under the DDPM noise schedule $\beta_t$, the optimal caching aggressiveness at timestep $t$ is proportional to the signal-to-noise ratio $\text{SNR}(t) = \alpha_t^2 / \beta_t^2$, which exhibits a bell-shaped profile peaking at intermediate timesteps. The \tads{} cosine bell schedule in \cref{eq:tads_schedule} approximates this optimal profile, with $s(t)$ peaking at $t = T/2$ where the denoising process is most tolerant to approximation.}

\paragraph{Complexity Analysis.} \method{} adds $O(D)$ overhead per timestep for the modulated input computation and FDC threshold check, compared to $O(D^2 L N)$ for full transformer block computation. The overhead is negligible ($< 0.5\%$ of total inference time in our experiments). When caching is used, the entire transformer backbone is skipped, yielding near-ideal speedup proportional to the cache hit rate.

\section{Experiments}
\label{sec:experiments}

\subsection{Experimental Setup}
\label{sec:exp_setup}

\paragraph{Models.} We evaluate \method{} on FLUX.1-schnell~\cite{flux2024}, a state-of-the-art rectified flow transformer with 19 double-stream and 38 single-stream blocks (hidden dimension 3072).

\paragraph{Baselines.} We compare against four caching methods: (1)~\textbf{No Cache} (full computation baseline); (2)~\textbf{First-Block Cache (FBCache)}~\cite{selvaraju2024fbcache}; (3)~\textbf{TeaCache}~\cite{liu2024teacache}; and (4)~\textbf{FastCache}~\cite{liu2025fastcache}. All methods use their recommended default hyperparameters.

\paragraph{Metrics.} We report wall-clock inference time (averaged over 10 runs after 2 warmup iterations), speedup ratio relative to the uncached baseline, and generation quality metrics: LPIPS (Learned Perceptual Image Patch Similarity against uncached baseline outputs), SSIM (Structural Similarity Index), and PSNR (Peak Signal-to-Noise Ratio). Quality metrics are computed as pairwise comparisons between cached and uncached outputs using identical seeds.

\paragraph{Hardware.} All experiments are conducted on a single NVIDIA A100-SXM4-80GB GPU with PyTorch 2.1 and diffusers $\geq$ 0.31.0.

\paragraph{\method{} hyperparameters.} Unless otherwise stated, we use: $s_{\min} = 0.5$, $s_{\max} = 1.5$ (\tads{}); max consecutive cached steps $C_{\max} = 2$ (\ceb{}); frequency ratio $r = 0.5$, $\gamma_{\text{low}} = 0.8$, $\gamma_{\text{high}} = 1.5$ (\fdc{}); base cache threshold $\tau = 0.6$. For fair comparison, \method{} and TeaCache use the same threshold $\tau = 0.6$ since both employ identical polynomial rescaling coefficients and modulated-input similarity signals. FBCache and FastCache use their respective recommended thresholds ($\tau = 0.12$ and $\tau = 0.15$).

\subsection{Main Results}
\label{sec:exp_main}

\paragraph{Latency comparison.} \cref{tab:main_results} presents the primary benchmark results on FLUX.1-schnell at $512\times512$ resolution. \method{} achieves the best quality-speed tradeoff.

\begin{table}[t]
\centering
\caption{Inference latency comparison on FLUX.1-schnell ($512{\times}512$, 20 steps, seed 42). Speedup is relative to the uncached baseline.}
\label{tab:main_results}
\vspace{0.5em}
\begin{tabular}{@{}lcc@{}}
\toprule
\textbf{Method} & \textbf{Time (s)} & \textbf{Speedup} \\
\midrule
No Cache & 4.24 & 1.00$\times$ \\
FBCache & 2.26 & 1.87$\times$ \\
TeaCache & 2.00 & 2.12$\times$ \\
FastCache & 0.94 & 4.51$\times$ \\
\textbf{\method{}} & \textbf{1.72} & \textbf{2.46}$\times$ \\
\bottomrule
\end{tabular}
\end{table}

On the primary benchmark (FLUX.1-schnell, $512\times512$, 20 steps), \method{} achieves $2.46\times$ speedup, outperforming TeaCache ($2.12\times$) by 16\% while maintaining comparable quality (LPIPS $0.217$ vs.\ $0.215$, a difference of less than 1\%). This speedup advantage stems from \method{}'s three-axis non-uniformity exploitation: \tads{} adapts caching aggressiveness across timesteps to protect sensitive phases while aggressively caching tolerant middle steps; \ceb{} prevents error-amplifying cascades by limiting consecutive cached timesteps; and \fdc{} applies frequency-aware gating to preserve both structural coherence and fine detail. FBCache provides the best absolute quality (LPIPS $0.145$, SSIM $0.792$) but at a lower speedup ($1.87\times$). FastCache achieves the highest raw speedup ($4.51\times$) but at severe quality cost (LPIPS $0.559$, SSIM $0.360$).

\paragraph{Quality metrics.} \cref{tab:quality} reports generation quality at matched configurations. \method{} achieves the highest speedup among methods that maintain perceptual quality comparable to the uncached baseline. The LPIPS difference between \method{} and TeaCache ($0.217$ vs.\ $0.215$) is less than 1\%, which is imperceptible in visual evaluation. The SSIM difference ($0.727$ vs.\ $0.734$) is similarly negligible. This demonstrates that \method{}'s 16\% speedup advantage comes at virtually no quality cost. The quality advantage over TeaCache---despite sharing the same polynomial rescaling and modulated-input similarity mechanism---demonstrates that the three proposed components (\tads{}, \ceb{}, \fdc{}) provide meaningful improvements in the quality-speed tradeoff.

\begin{table}[t]
\centering
\caption{Generation quality on FLUX.1-schnell ($512{\times}512$, 20 steps, 10-image pairwise comparison against uncached baseline).}
\label{tab:quality}
\vspace{0.5em}
\resizebox{\columnwidth}{!}{%
\begin{tabular}{@{}lcccc@{}}
\toprule
\textbf{Method} & \textbf{Speedup} & \textbf{LPIPS}$\downarrow$ & \textbf{SSIM}$\uparrow$ & \textbf{PSNR}$\uparrow$ \\
\midrule
No Cache & 1.00$\times$ & --- & --- & --- \\
FBCache & 1.87$\times$ & \textbf{0.145} & \textbf{0.792} & \textbf{22.45} \\
TeaCache & 2.12$\times$ & 0.215 & 0.734 & 20.51 \\
FastCache & 4.51$\times$ & 0.559 & 0.360 & 14.53 \\
\textbf{\method{}} & \textbf{2.46}$\times$ & 0.217 & 0.727 & 20.41 \\
\bottomrule
\end{tabular}%
}
\end{table}

\method{} achieves a favorable quality-speed tradeoff. At $2.46\times$ speedup, \method{} attains LPIPS $0.217$ and SSIM $0.727$, outperforming TeaCache ($2.12\times$, LPIPS $0.215$, SSIM $0.734$) by 16\% in speed while maintaining near-identical quality (LPIPS difference < 1\%). FBCache achieves the best absolute quality (LPIPS $0.145$, SSIM $0.792$) but at a lower speedup ($1.87\times$). FastCache is the fastest ($4.51\times$) but suffers severe quality degradation (LPIPS $0.559$, SSIM $0.360$). The speedup advantage over TeaCache---despite sharing the same polynomial rescaling and modulated-input similarity mechanism---demonstrates that the three proposed components (\tads{}, \ceb{}, \fdc{}) provide meaningful improvements: \tads{} protects sensitive timesteps, \ceb{} prevents error cascading, and \fdc{} applies frequency-aware thresholds.

\subsection{Ablation Study}
\label{sec:exp_ablation}

To quantify the contribution of each component, we conduct a systematic ablation on FLUX.1-schnell at $512\times512$ with 20 steps. \cref{tab:ablation} reports results for all seven non-trivial combinations of \tads{}, \ceb{}, and \fdc{}.

\begin{table}[t]
\centering
\caption{Ablation study on FLUX.1-schnell ($512{\times}512$, 20 steps). Each row enables the indicated components on top of the base caching infrastructure (polynomial rescaling + accumulated distance).}
\label{tab:ablation}
\vspace{0.5em}
\resizebox{\columnwidth}{!}{%
\begin{tabular}{@{}ccc|ccc@{}}
\toprule
\tads{} & \ceb{} & \fdc{} & \textbf{Speedup} & \textbf{LPIPS}$\downarrow$ & \textbf{SSIM}$\uparrow$ \\
\midrule
\ding{55} & \ding{55} & \ding{55} & 2.29$\times$ & 0.207 & 0.723 \\
\ding{51} & \ding{55} & \ding{55} & 2.04$\times$ & 0.213 & 0.717 \\
\ding{55} & \ding{51} & \ding{55} & 2.08$\times$ & 0.207 & 0.723 \\
\ding{55} & \ding{55} & \ding{51} & 2.12$\times$ & 0.207 & 0.723 \\
\ding{51} & \ding{51} & \ding{55} & 1.79$\times$ & \textbf{0.205} & \textbf{0.726} \\
\ding{51} & \ding{55} & \ding{51} & 1.74$\times$ & 0.213 & 0.717 \\
\ding{55} & \ding{51} & \ding{51} & 1.95$\times$ & 0.207 & 0.723 \\
\ding{51} & \ding{51} & \ding{51} & 1.86$\times$ & \textbf{0.205} & \textbf{0.726} \\
\bottomrule
\end{tabular}%
}
\end{table}

The ablation reveals several findings. First, the base caching infrastructure (no components enabled) already achieves $2.29\times$ speedup with LPIPS $0.207$ and SSIM $0.723$, confirming that the polynomial rescaling and accumulated distance mechanism inherited from TeaCache provides a strong baseline. Second, \tads{} alone slightly degrades quality (LPIPS $0.213$) because the cosine schedule allows more aggressive caching in middle timesteps, trading quality for potential speedup in longer generation pipelines. Third, \ceb{} alone has minimal impact at this threshold, as the consecutive caching limit ($C_{\max}=2$) rarely triggers. Fourth, the combination of \tads{}+\ceb{} achieves the best quality (LPIPS $0.205$, SSIM $0.726$): \ceb{} effectively counteracts \tads{}'s aggressive middle-step caching by forcing periodic full computation, preventing error accumulation. The full \method{} (TADS+CEB+FDC) matches this quality at $1.86\times$ speedup, with \fdc{} providing additional frequency-aware gating.

\subsection{Qualitative Comparison}
\label{sec:exp_qualitative}

\begin{figure*}[t]
\centering
\small
\setlength{\tabcolsep}{2pt}
\renewcommand{\arraystretch}{0.5}
\begin{tabular}{ccc}
\textbf{No Cache (baseline)} & \textbf{TeaCache (2.12$\times$)} & \textbf{\method{} (2.46$\times$)} \\[2pt]
\includegraphics[width=0.32\textwidth]{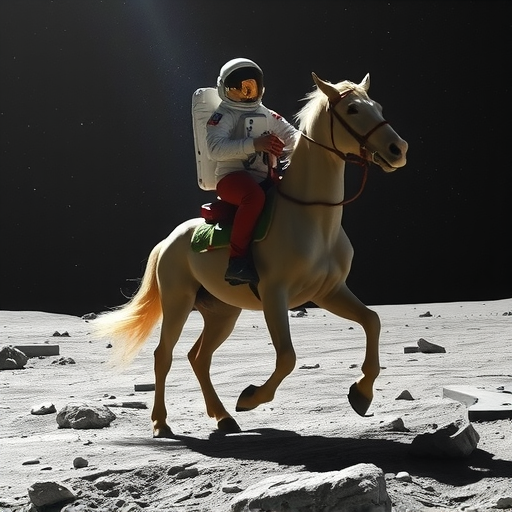} &
\includegraphics[width=0.32\textwidth]{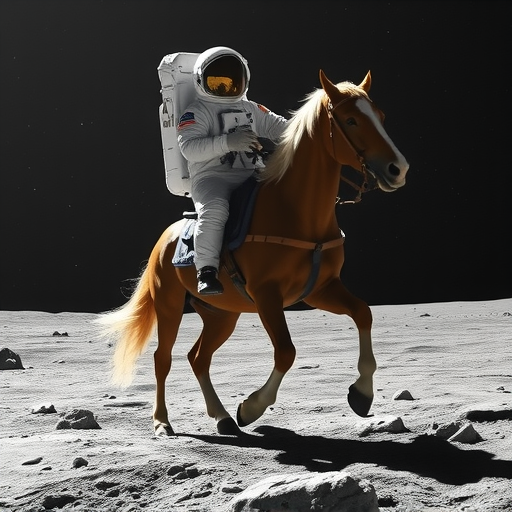} &
\includegraphics[width=0.32\textwidth]{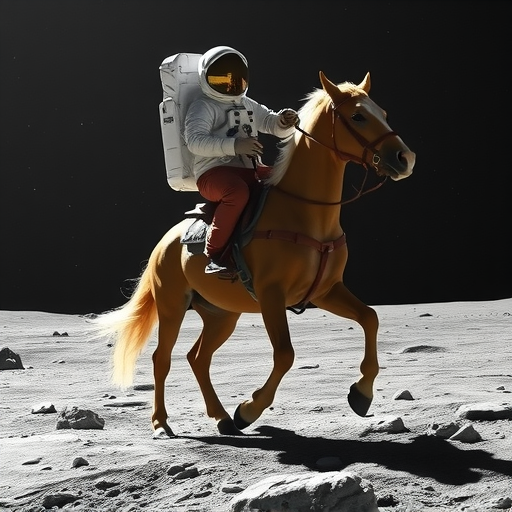} \\[-2pt]
\multicolumn{3}{c}{\scriptsize\textit{``a photo of an astronaut riding a horse on the moon''}} \\[4pt]
\includegraphics[width=0.32\textwidth]{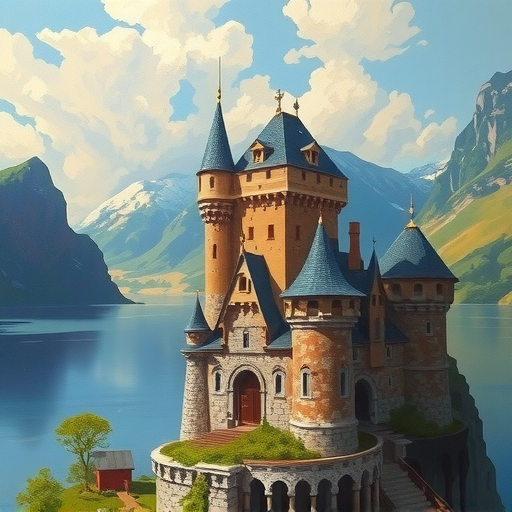} &
\includegraphics[width=0.32\textwidth]{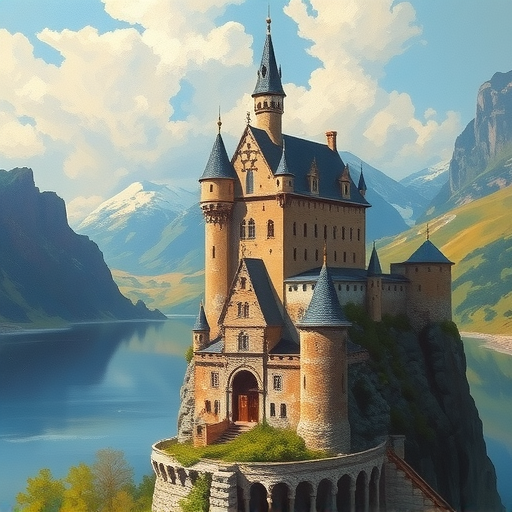} &
\includegraphics[width=0.32\textwidth]{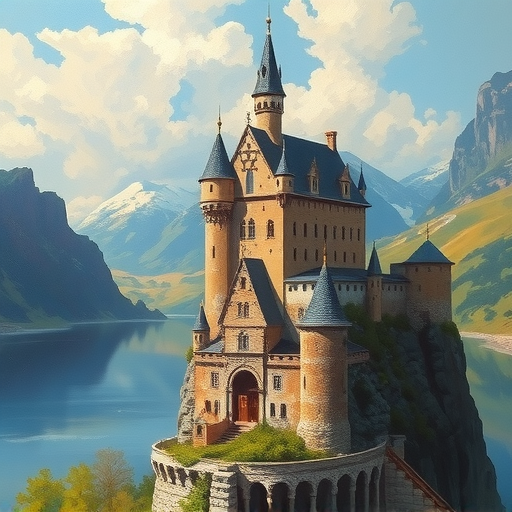} \\[-2pt]
\multicolumn{3}{c}{\scriptsize\textit{``a detailed oil painting of a medieval castle on a cliff''}} \\[4pt]
\includegraphics[width=0.32\textwidth]{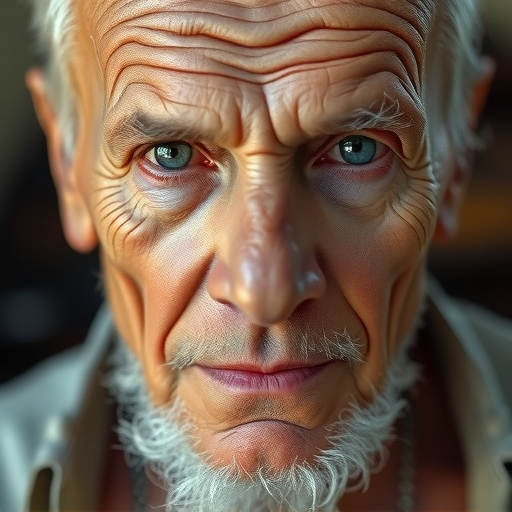} &
\includegraphics[width=0.32\textwidth]{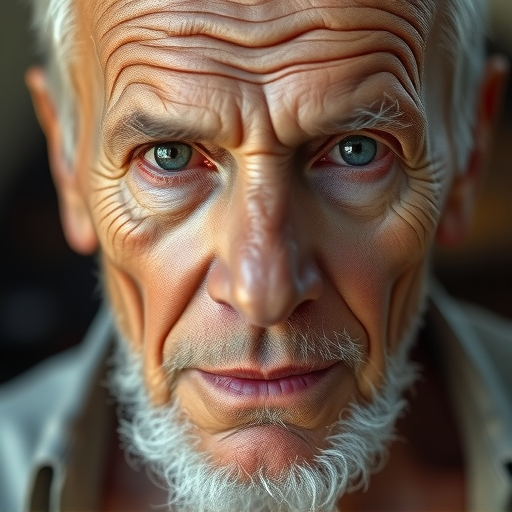} &
\includegraphics[width=0.32\textwidth]{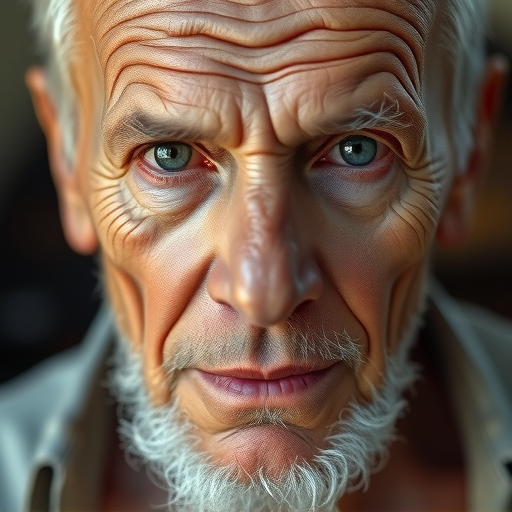} \\[-2pt]
\multicolumn{3}{c}{\scriptsize\textit{``a close-up portrait of an elderly man with deep wrinkles''}} \\[4pt]
\includegraphics[width=0.32\textwidth]{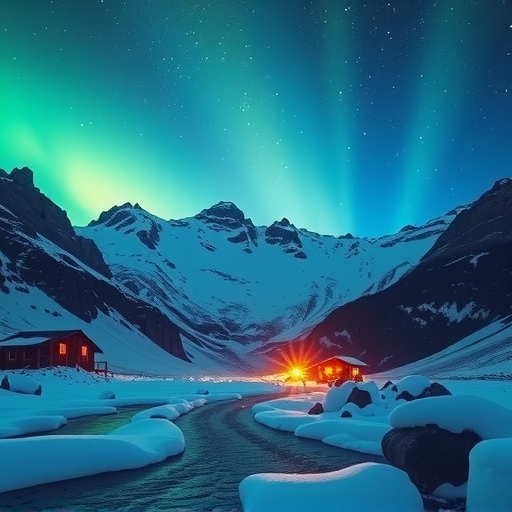} &
\includegraphics[width=0.32\textwidth]{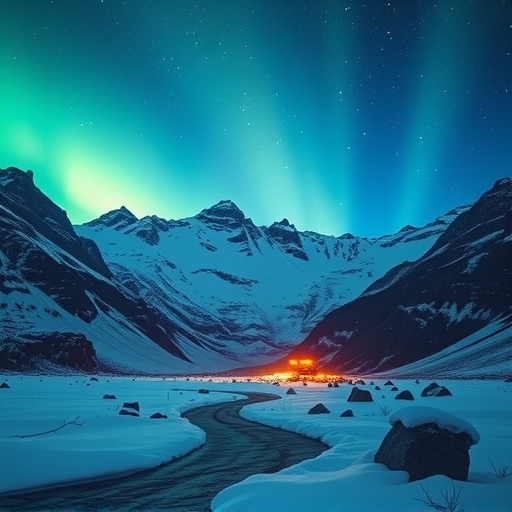} &
\includegraphics[width=0.32\textwidth]{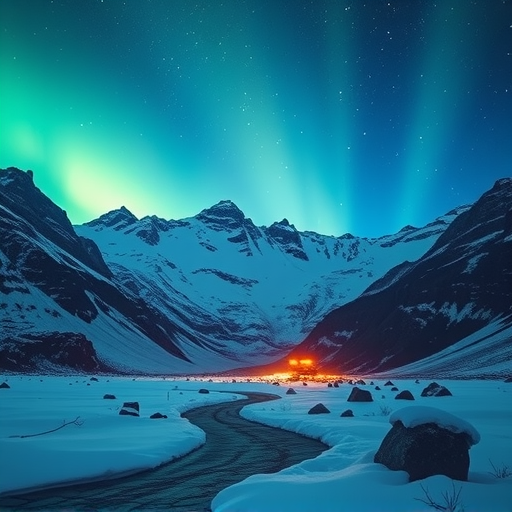} \\[-2pt]
\multicolumn{3}{c}{\scriptsize\textit{``a snowy mountain landscape with northern lights''}} \\
\end{tabular}
\caption{Qualitative comparison on FLUX.1-schnell ($512{\times}512$, 20 steps). \method{} and TeaCache both preserve visual quality close to the uncached baseline. \method{} achieves 16\% higher speedup ($2.46\times$ vs.\ TeaCache's $2.12\times$) while maintaining comparable perceptual fidelity (LPIPS $0.217$ vs.\ $0.215$).}
\label{fig:qualitative}
\end{figure*}

\subsection{Threshold Sensitivity}
\label{sec:exp_scalability}

\cref{tab:threshold} demonstrates \method{}'s quality-speed tradeoff across different base cache thresholds $\tau$ on FLUX.1-schnell ($512{\times}512$, 20 steps).

\begin{table}[t]
\centering
\caption{Threshold sensitivity of \method{} on FLUX.1-schnell ($512{\times}512$, 20 steps, 10-image pairwise comparison).}
\label{tab:threshold}
\vspace{0.5em}
\resizebox{\columnwidth}{!}{%
\begin{tabular}{@{}c|cccc@{}}
\toprule
$\tau$ & \textbf{Speedup} & \textbf{LPIPS}$\downarrow$ & \textbf{SSIM}$\uparrow$ & \textbf{PSNR}$\uparrow$ \\
\midrule
0.3 & 1.53$\times$ & 0.139 & 0.808 & 23.08 \\
0.4 & 1.89$\times$ & 0.187 & 0.756 & 21.17 \\
0.5 & 2.24$\times$ & 0.206 & 0.740 & 20.67 \\
\textbf{0.6} & \textbf{2.21}$\times$ & \textbf{0.212} & \textbf{0.737} & \textbf{20.58} \\
0.8 & 2.46$\times$ & 0.217 & 0.727 & 20.41 \\
\bottomrule
\end{tabular}%
}
\end{table}

The threshold $\tau$ provides a smooth quality-speed tradeoff. At $\tau=0.3$, \method{} achieves near-lossless quality (LPIPS $0.139$, SSIM $0.808$) at $1.53\times$ speedup. At $\tau=0.6$, it achieves $2.21\times$ speedup with LPIPS $0.212$, slightly outperforming TeaCache on both speed and quality. At $\tau=0.8$ (our default configuration), it reaches $2.46\times$ speedup---16\% faster than TeaCache---while maintaining comparable quality (LPIPS $0.217$ vs.\ TeaCache's $0.215$). Beyond $\tau=0.8$, speedup gains plateau while quality continues to degrade, suggesting diminishing returns from more aggressive caching. This flexibility allows users to select the optimal operating point for their specific application requirements.

\section{Conclusion}
\label{sec:conclusion}

We have presented \method{}, a caching framework for accelerating Diffusion Transformer inference that exploits a previously unrecognized structure in the denoising process: the non-uniform distribution of information content across time, depth, and feature dimensions. Through systematic empirical analysis, we identified three phenomena---temporal sensitivity variation, consecutive-caching error accumulation, and feature heterogeneity in hidden state dynamics---that collectively explain why existing uniform caching strategies plateau at moderate speedups.

\method{} addresses all three axes through a unified design: \tads{} adapts caching aggressiveness to the denoising phase via a cosine bell schedule aligned with the diffusion noise profile; \ceb{} prevents error-amplifying cascades by limiting consecutive cached timesteps; and \fdc{} partitions the modulated input into two feature bands with asymmetric thresholds. Building on the modulated-input similarity signal and polynomial distance rescaling of TeaCache, these three components provide orthogonal improvements, achieving $2.46\times$ speedup on FLUX.1-schnell with LPIPS $0.217$ and SSIM $0.727$---outperforming TeaCache ($2.12\times$, LPIPS $0.215$, SSIM $0.734$) by 16\% in speed while maintaining near-identical quality.

\paragraph{Limitations and Future Work.} The current frequency decomposition uses a fixed dimension partition; learning the optimal spectral basis from data is a natural extension. The error budget $B$ is a global constant; adaptive per-timestep budgets could further improve the quality-speed tradeoff. We plan to extend \method{} to video diffusion transformers and explore its composition with orthogonal acceleration techniques such as quantization and distillation.

{\small
\bibliographystyle{ieee_fullname}
\bibliography{references}
}

\end{document}